\documentclass[fleqn,10pt]{wlscirep}
\usepackage[utf8]{inputenc}
\usepackage[T1]{fontenc}

\usepackage{graphicx}%
\usepackage{multirow}%
\usepackage{amsmath,amssymb,amsfonts}%
\usepackage{amsthm}%
\usepackage{mathrsfs}%
\usepackage[title]{appendix}%
\usepackage{xcolor}%
\usepackage{textcomp}%
\usepackage{manyfoot}%
\usepackage{booktabs}%
\usepackage{algorithm}%
\usepackage{algorithmicx}%
\usepackage{algpseudocode}%
\usepackage{listings}%

\usepackage[acronym]{glossaries}
\usepackage{xspace} 
\usepackage{siunitx}

\usepackage{tikz}
\usepackage{pgfplots}
\pgfplotsset{compat=1.5} 

\usepackage{caption}
\usepackage{subcaption}

\newacronym{pw}{PW}{powered wheelchair}
\newacronym{ipw}{IPW}{intelligent powered wheelchair}

\newglossaryentry{slam}
{
    name= SLAM,
    description={simultaneous localization and mapping}
}

\newcommand{\bm}[1]{\boldmath{#1}}

\newcommand{\xState}{\boldsymbol{x}}
\newcommand{\xAngle}{\Theta}
\newcommand{\xVel}{v}
\newcommand{\angleDes}{\xAngle_\mathrm{des}}

\newcommand{\uRot}{u_{\mathrm{rot}}}
\newcommand{\uTrans}{u_{\mathrm{trans}}}
\newcommand{\uHRot}{u_{\mathrm{H,rot}}}
\newcommand{\uHTrans}{u_{\mathrm{H,trans}}}
\newcommand{\uARot}{u_{\mathrm{A,rot}}}

\newcommand{\FL}{F_{\mathrm{L}}}
\newcommand{\FR}{F_{\mathrm{R}}}
\newcommand{\rotFac}{a_\mathrm{rot}}
\newcommand{\transFac}{a_\mathrm{trans}}

\newcommand{\varFac}{\lambda}
\newcommand{\icp}{k_{\mathrm{cp}}}
\newcommand{\ila}{k_{\mathrm{la}}}
\newcommand{\lap}{p_\mathrm{la}}
\newcommand{\lapX}{p_\mathrm{la,x}}
\newcommand{\lapY}{p_\mathrm{la,y}}
\newcommand{\lad}{d_{\mathrm{la}}}
\newcommand{\assFac}{\gamma}

\newcommand{\xMean}{\boldsymbol{\bar{x}}}   

\newcommand{\cov}[1]{\mathrm{cov}(#1)}

\newcommand{\mNosup}{\textit{noSup}\xspace}
\newcommand{\mHighvar}{\textit{highVar}\xspace}
\newcommand{\mLowvar}{\textit{lowVar}\xspace}

\title{Locomotion Variability and User Experience in Smart Wheelchair Human--Robot Interaction}

\author[1,*]{Sean Kille}
\author[2]{Adina M. Panchea}
\author[1]{Balint Varga}
\author[1]{Sören Hohmann}
\affil[1]{Karlsruhe Institute of Technology (KIT), Institute of Control Systems (IRS), Karlsruhe, 76131, Germany}
\affil[2]{Université de Sherbrooke, Interdisciplinary Institute for Technological Innovation (3IT), Sherbrooke, J1N 3C6, Canada}

\affil[*]{sean.kille@kit.edu}

\begin{abstract}
Human movement is inherently variable, with variability structured according to task relevance: movements are typically more consistent at task-critical points and more flexible elsewhere. In human--robot interaction (HRI), however, model-based assistance strategies commonly assume deterministic human behavior and suppress such variability, potentially altering how interactions are experienced and lowering sense of agency. While movement variability is increasingly recognized as functionally meaningful, its deliberate preservation in assisted interaction---and its consequences for user experience---remain underexplored.

In this paper, we empirically investigate how different assistance strategies shape human movement variability, task performance, and subjective interaction experience in a shared control setting. 
We introduce an autonomy-supportive shared control strategy that preserves users' natural movement structure.
This approach is evaluated in a user study in which participants push an intelligent powered wheelchair under three conditions: no assistance, conventional variability-reducing assistance, and variability-preserving assistance.

While task-relevant performance remained comparable across assisted modes, preserving natural movement variability led to more favorable interaction experiences. In particular, participants reported significantly higher perceived agency compared to conventional assistance and highest perceived usefulness. These findings suggest that variability-aware assistance can support both performance and user autonomy in physical human–robot collaboration. More broadly, the results highlight the importance of designing assistive robotic systems that respect the embodied structure of human movement rather than treating variability as noise to be neglected or eliminated.
\end{abstract}
\begin{document}

\flushbottom
\maketitle
%
%
\thispagestyle{empty}

\section*{Introduction}

Robotic systems spans a wide range of applications, many of which pursue full automation, particularly in highly repetitive tasks within structured environments. However, numerous scenarios benefit significantly from human--robot interaction (HRI) designed as shared control systems, where human and robot collaborate in a physically coupled manner. Such settings are prevalent in manufacturing---especially for individualized or small-batch production~\cite{Alt.2024}---and in unstructured environments, e.g., building \mbox{(re-)construction} or circular architecture. Moreover, automation plays a crucial role in assistive contexts, including rehabilitation and physiotherapy \cite{Schneider.2024, Schneider.2025}, where it supports individuals in regaining movement and strength, as well as in daily tasks such as walking assistance. 
With a trend towards human-computer integration \cite{Mueller.2022} and scenarios in which human and automation aim for symbiotic interaction, the quality of the interaction experience becomes increasingly crucial \cite{Inga.2023}.

To design effective assistance mechanisms, it is essential to first understand natural human behavior. Studies have consistently shown that repetitive human movements are inherently variable; they are not deterministic but stochastic in nature~\cite{Abend.1982, Harris.1998}. Importantly, human movement variability follows a structured pattern: lower variability in task-relevant regions and higher variability in task-irrelevant areas---a phenomenon known as task-relevant variability~\cite{Todorov.2002} and widely described by the linear-quadratic sensorimotor (LQS) model~\cite{Todorov.2005}. By incorporating both additive and multiplicative noise processes in the human action-perception loop, the LQS model extends the classical deterministic linear-quadratic framework, resulting in a more accurate representation of the human mean movement behavior and variability~\cite{Karg.2023, Mitrovic.2011}.

Most model-based control approaches in HRI largely disregard human movement variability. Humans are commonly modeled as deterministic impe\-dances \cite{Ficuciello.2015, Dong.2020} or as deterministic optimal controllers within differential games~\cite{Flad.2017, Varga.2024}, neglecting their inherent stochasticity. This results in an overall reduction of variability during interaction. 

While some literature addresses the natural stochasticity of human movements, initial efforts treat uncertainty to be minimized or compensated. For example, Medina et al.~\cite{Medina.2015} propose an optimal assistance strategy that adapts the control cost function based on behavioral or sensing uncertainty, but without the aim of retaining natural variability. Similarly, Gribovskaya et al.~\cite{Gribovskaya.2011} adjust controller impedance to counteract variability, without preserving it.

Several more recent studies have emphasized the functional role of variability in HRI, particularly in the context of motor learning and rehabilitation. Fitzsimons et al.~\cite{Fitzsimons.2020} introduced a task-based hybrid shared control framework for motor training, where robotic assistance is recomputed at each step to allow natural movement variability while ensuring task success. 
Similarly, Özen et al.~\cite{Ozen.2021} demonstrated that intentionally promoting motor variability during robotic assistance enhances motor learning in dynamic tasks. 
Variability can even be intentionally introduced through external perturbations to enhance motor learning and muscle capacity~\cite{Rubino.2024}.
Together, these works highlight that preserving variability can play a beneficial role in training and recovery.

Beyond motor learning applications, distribution-based approaches have 
increasingly been applied to HRI as a means of representing and leveraging variability. Fitzsimons and Murphey~\cite{Fitzsimons.2022} introduced an ergodic shared control framework that closes the loop based on information distributions encoded in human motion, thereby allowing robots to align with stochastic aspects of human behavior. Similarly, Maeda
et al.~\cite{Maeda.2017} proposed Probabilistic Movement Primitives (ProMPs) as a means to represent and reproduce distributions of human trajectories in collaborative tasks. 
These approaches underline the growing recognition that variability is not simply noise, but carries meaningful structure that can be exploited for more adaptive interaction.
While this prior work demonstrates how variability can be preserved and leveraged in HRI, it primarily focuses on algorithmic control strategies or motor learning outcomes. 
Building on this foundation, our earlier simulations demonstrated that explicitly incorporating human noise processes into model-based control designs can enhance joint task performance~\cite{Kille.2024}.
However, little is known about how variability-preserving designs affect the human experience of interaction, particularly in shared control settings where user autonomy and system assistance must be balanced. 
Moreover, the effect of preserving variability on usability, acceptance and perceived agency in physically coupled human-machine systems remains largely unexplored.
This study addresses this gap by empirically examining how preserving natural locomotion variability influences both objective performance and subjective experience in a powered wheelchair task.

To the best of our knowledge, no prior work has explicitly sought to preserve natural human variability in HRI systems and has explicitly analyzed the effect that variability adaption has on the user experience. 
    In this paper, we therefore focus on the design and evaluation of a variability-preserving shared control framework. Our central hypothesis is, 
that restricting a human's inherent variability negatively affects the interaction experience. 
To test this hypothesis, a user study was carried out, in which participants interacted with an \gls{ipw}~\cite{Panchea.2022} under three different automation assistance modes. 
Among these, we introduce a novel assistance mode, which integrates human variability patterns into a shared control framework. This new mode selectively allows variability in task-irrelevant regions while providing guidance near task-relevant points. 
By analyzing both objective performance and subjective experience, we provide empirical evidence on the role of natural locomotion variability in shared control and its implications for the design of human-centric HRI systems.

The main contributions of this study are as follows:

\begin{itemize}
    \item The design of a shared control framework and a novel assistance mode that explicitly incorporates human natural variability patterns, preserving variability in task-irrelevant areas while maintaining assistance in task-relevant regions.
    \item An empirical evaluation through a user study with an IPW under three different automation modes, providing systematic evidence on the effects of preserving or restricting variability on both task performance and user experience.
    \item Objective results demonstrating that preserving variability does not compromise task-relevant performance. Subjective results revealing the relationship between variability-preserving assistance and user experience, particularly highlighting the role of perceived agency and ease of use.
\end{itemize}

\textbf{Terminology Clarification.} In this work, we use the term \textit{shared control} to refer to a continuous blending of human and automation inputs~\cite{Flemisch.2016}. 
Unlike switching-based approaches, the system does not override or replace the user's commands but instead modulates the level of assistance in relation to task relevance. 
We define \textit{variability} as the trial-to-trial differences in the human movement trajectories, particularly in the spatial domain. 
The distinction between \textit{task-relevant} and \textit{task-irrelevant areas} is operationalized with respect to spatial proximity to key trajectory segments, such as the start and goal regions. 
Task-relevant areas are those where accurate performance is critical to achieving the goal, while task-irrelevant areas allow more freedom in movement.

\section{System Design}

\subsection{Intelligent powered wheelchair as a shared control system}\label{sec:ipw}

\begin{figure}[t]
  \begin{center}
  \includegraphics[width=0.35\textwidth]{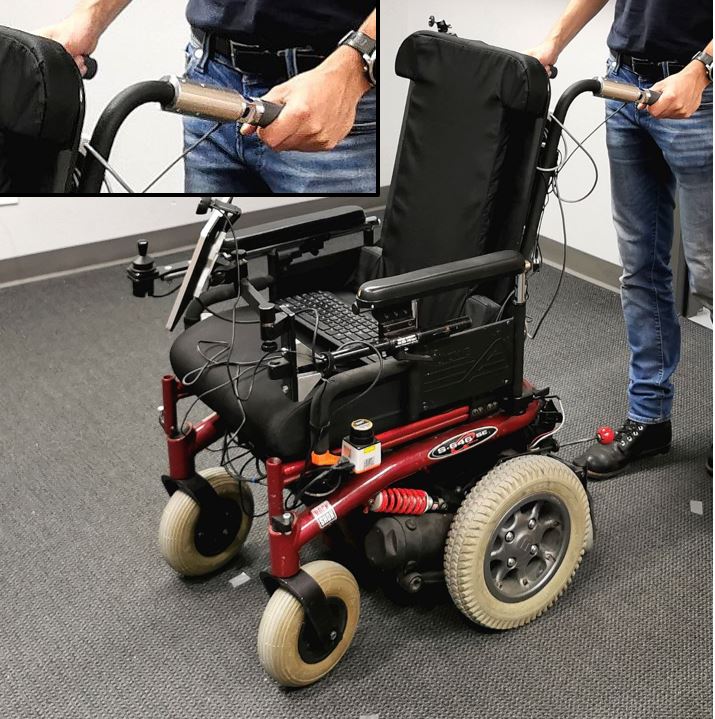}\\
  \caption{Intelligent powered wheelchair being pushed by a human user, with force sensors mounted on the rear handles to measure input forces.}\label{fig:ipw}
  \end{center}
\end{figure}

The \gls{ipw} introduced in Panchea et al.~\cite{Panchea.2022}, depicted in Fig.~\ref{fig:ipw}, and previously used as a human-only system in Kille et al.~\cite{Kille.2025}, is adapted to be used as a human-machine shared control system. It is modeled as a differential drive platform actuated by two electric motors. The system state is comprised of the position $(p_x, p_y)$, orientation $\xAngle$, and translational velocity $\xVel$ of the wheelchair, i.e., $\xState = [p_x \quad p_y \quad \xAngle \quad \xVel]^\intercal$. 
Control inputs are a translational velocity $\uTrans$ and rotational velocity $\uRot$, which are executed directly by the motor controllers. In this study, the translational set velocity is determined by the human translational push and pull $\uHTrans$; rotational set velocities are to be set by the human ($\uHRot$) and automation ($\uARot$) rotation collectively: 

\begin{align}
\uTrans & = \uHTrans, \\
\uRot & = \uARot + \uHRot.
\end{align}

We assume no-slip kinematics; a dynamic model is not required for the assistance law used in this study.

To imitate a conventional, non-powered wheelchair being pushed by a human, 1-D force sensors are attached to the two push handles, measuring the forward push forces on the left and right handles, $\FL$ and $\FR$. These forces are translated into the human input using scaling factors $\transFac$ and $\rotFac$ as follows:
\begin{align}
\uHTrans & = \transFac ( \FL + \FR) \quad \mathrm{and} \\
\uHRot & = \rotFac (\FL - \FR).
\end{align}

The force sensors are calibrated such that a maximum overall force of $\FL + \FR = \SI{50}{N}$ on both handles leads to the maximum translational input of $\uHTrans = 1$ (corresponding to a translational velocity of $\SI{0.75}{m/s}$) and a delta force of $\FL - \FR = \SI{40}{N}$ leads to the maximum rotational input of $\uHTrans = 1$ (corresponding to a rotational velocity of $\SI{1.0}{rad/s}$). This results in scaling factors of $\transFac = 0.02$ and $\rotFac = 0.025$.

Localization of the \gls{ipw}'s position and orientation is achieved using a Simultaneous Localization and Mapping (\gls{slam}) approach based on the RTAB-Map library by Labbé and Michaud~\cite{Labbe.2019}. This library utilizes LIDAR sensor data combined with odometry information from the wheel encoders to construct and update the environment map, as illustrated in Fig.~\ref{fig:lab}.

This technical setup not only enables shared control between human and system, but also allows us to characterize and later adapt to the user's natural movement patterns, which are introduced in the following section.

\subsection{Human natural movement}\label{sec:humanMovement}

A key aspect of this study is capturing the natural movement variability of participants when interacting with the IPW without automation support. 
This baseline behavior serves both to tailor the assistance modes and to 
quantify inherent human variability.

To capture participants natural movement behavior and to assess their natural variability, participants interact with the \gls{ipw} with the \mNosup mode, i.e., without automation support, $\uARot = 0$. Each subject performs $15$ repetitions of the task (see Sec.~\ref{sec:task}) under this condition.

To standardize the trajectory representation and eliminate velocity information, each trajectory is cropped to the length between the given start point and endpoint and resampled using spline interpolation to obtain $300$ evenly spaced data points along its path. Based on the forward repetitions, a natural mean path $\xMean_\mathrm{noSup}(k)$ is computed individually for each subject, serving as a reference for subsequent automation modes and variability analysis. The index $k$ denotes the spatial progression along the mean path.

\subsection{Automation assistance modes} \label{sec:automation}

Two assistance modes are implemented in the shared control system: \mHighvar and \mLowvar. In \mHighvar mode, assistance is reduced in task-irrelevant areas, while in \mLowvar mode, assistance is applied continuously throughout the entire interaction.

The assistance is realized using a pure pursuit path-following approach, guiding the \gls{ipw} along a previously identified natural mean path $\xMean_\mathrm{noSup} $ of each individual (which is identified as described in Sec.~\ref{sec:humanMovement}) by providing an additional rotational control force $\uARot$. Both modes require the following inputs: the current \gls{ipw} position and orientation, as contained in the state $\xState$, the mean path $\xMean_\mathrm{noSup}$, a lookahead distance $\lad$, an assistance gain factor $\assFac$, and a variability factor function $\varFac(k)$.
 
At runtime, let
$
\icp = \arg\min_{k}\|\xMean_\mathrm{noSup}(k)-[p_x\;\;p_y]^\top\|
$
be the closest path index, and define the index of the look-ahead point as
$
\ila = \icp + \lad
$. 
This lets us find the look-ahead point as
$\lap = \xMean_\mathrm{noSup}(\ila).$

Define the desired heading toward the look-ahead point as
\begin{equation}
\angleDes = \operatorname{atan}\!\big(\,\lapY-p_y,\;
\lapX-p_x\,\big).
\end{equation}
The heading error is $e_\theta=\angleDes-\xAngle$ (wrapped to $(-\pi,\pi]$).

Both modes share the same linear feedback structure with gain $\assFac>0$ and
a \emph{variability factor} $\varFac(k)\in[0,1]$:
\begin{equation}
u_{A,\mathrm{rot}} \;=\; \lambda\!\big(k_{\mathrm{la}}\big) \; \assFac \; e_\theta.
\label{eq:assist}
\end{equation}
Mode \mLowvar uses a constant factor
\begin{equation}
\lambda_{\mathrm{lowVar}}(k) \equiv 1,
\end{equation}
i.e., uniform assistance along the path.

\begin{figure}[tbp]
  \centering
  \resizebox{.5\columnwidth}{!}{
      \input{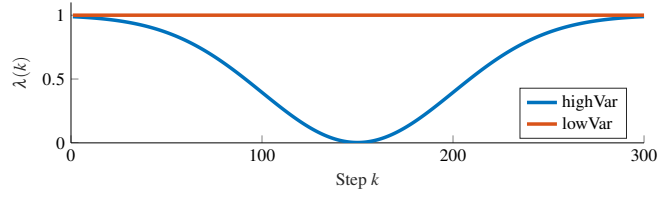}
      }
  \caption{Variability factor $\varFac$ for \mHighvar and \mLowvar modes along the spatial path index $k$. }
  \label{fig:varFac}
\end{figure}

Natural human movement patterns show bell-shaped variability patterns \cite{Abend.1982,Kille.2024a}, showing high variability in the task-irrelevant midsection between target points and a reduced variability at target points. In order to mirror these patterns, the variability factor in \mHighvar mode is modeled as an inverted bell-shaped curve, defined as a function over the spatial path index $k$, i.e., progressing along the mean path, as illustrated in Fig.~\ref{fig:varFac}. 
The factor is set to $\varFac = 1$ at the start and end of the path, decreasing to $\varFac = 0$ at the midpoint of the path. This design ensures that assistance is minimized in task-irrelevant areas, allowing for greater user control and variability, while being maximized in task-relevant areas to enhance precision and guidance.

The detailed algorithm used to compute the rotational assistance input is presented in Algorithm~\ref{alg:one}.

In this implementation, assistance was applied as an additive rotational correction $u_{A,\mathrm{rot}}$, while forward propulsion remained entirely under human control. This allows the assistance to blend its rotational input with that of the user without overriding it. 
The rationale was to isolate the influence of positional variability on the user experience: altering translational velocity would have changed workload and effort, thereby introducing additional confounding factors. 
Restricting automation to rotational corrections ensured that differences between modes reflected only the intended manipulation of variability.

\begin{algorithm}
\caption{Pure Pursuit Path-Following with Variability Modulation}\label{alg:one}
\begin{algorithmic}[1]

\Require 
\State $\xState=[p_x \quad p_y \quad \xAngle \quad \xVel]^\intercal$ \Comment{current state}
\State $\xMean_\mathrm{noSup}$ \Comment{natural mean path}
\State $\lad$ \Comment{lookahead distance}
\State $\assFac$ \Comment{assistance gain factor}
\State $\varFac(k)$ \Comment{variability factor function (see Fig.~\ref{fig:varFac})}

\Ensure $\uARot$ \Comment{rotational control input}

\Loop
    \State find $\icp \Leftarrow \arg\min_{k}\left\|\xMean_\mathrm{noSup}-[p_x\;\;p_y]^\top\right\|$
        \Comment{index of closest point on $\xMean_\mathrm{noSup}$ to current position}
    \State compute $\ila \Leftarrow \icp + \lad$
        \Comment{move forward along $\xMean_\mathrm{noSup}$ by $\lad$}
    \State compute $\lap \Leftarrow \xMean_\mathrm{noSup}(\ila)$
        \Comment{lookahead point}
    \State compute $\angleDes$
        \Comment{desired orientation from $\xState$ to $\lap$}
    \State compute $e_\theta \Leftarrow \angleDes - \xAngle$
        \Comment{heading error}
    \State compute $\uARot \Leftarrow \varFac(\ila)\;\assFac\; e_\theta$
        \Comment{rotational control input}
\EndLoop

\end{algorithmic}
\end{algorithm}

\section{Study Design}

This study analyses the influence of different automation strategies---particularly variability-preserving assistance---on performance metrics and user perceptions in tasks involving shared locomotion between humans and automated systems. 
To achieve this, we compare three distinct mode conditions as independent variables: one mode without assistance support (\mNosup), one mode providing continuous assistance (with the result of low task-irrelevant variability: \mLowvar), and one mode preserving high task-irrelevant variability (\mHighvar). 
The modes offer different implementations of an automation in a shared locomotion task. 
As a baseline condition, the \mNosup mode captures each participant's natural movement patterns and variability, as well as interaction experience without any assistance. This baseline provides the reference against which the effects of the assisted modes can be interpreted.

To structure our analysis, the following hypotheses are introduced in our study:

\begin{itemize}
\item \textit{H1}: The \mHighvar mode leads to an increased task-irrelevant variability compared to \mLowvar mode.
\item \textit{H2}: No observable difference in task-relevant performance exists between \mHighvar and \mLowvar mode.
\item \textit{H3}: The \mHighvar mode will lead to an increased sense of agency, perceived usefulness and ease of use compared to \mLowvar mode.
\end{itemize}

The first hypothesis \textit{H1} aims at validating whether our control modes work as intended, i.e., leading to variations in task-irrelevant variability. The influence of the control modes on objective task performance is analyzed with hypothesis \textit{H2}, while hypothesis \textit{H3} evaluates how variability preservation influences user experience. 

The objective and subjective measures, which constitute the dependent variables, are detailed in Sections~\ref{sec:objMeasures} and~\ref{sec:experience}, respectively.

\subsection{Task}\label{sec:task}

\begin{figure}
    \begin{center}
    \includegraphics[width=0.4\columnwidth]{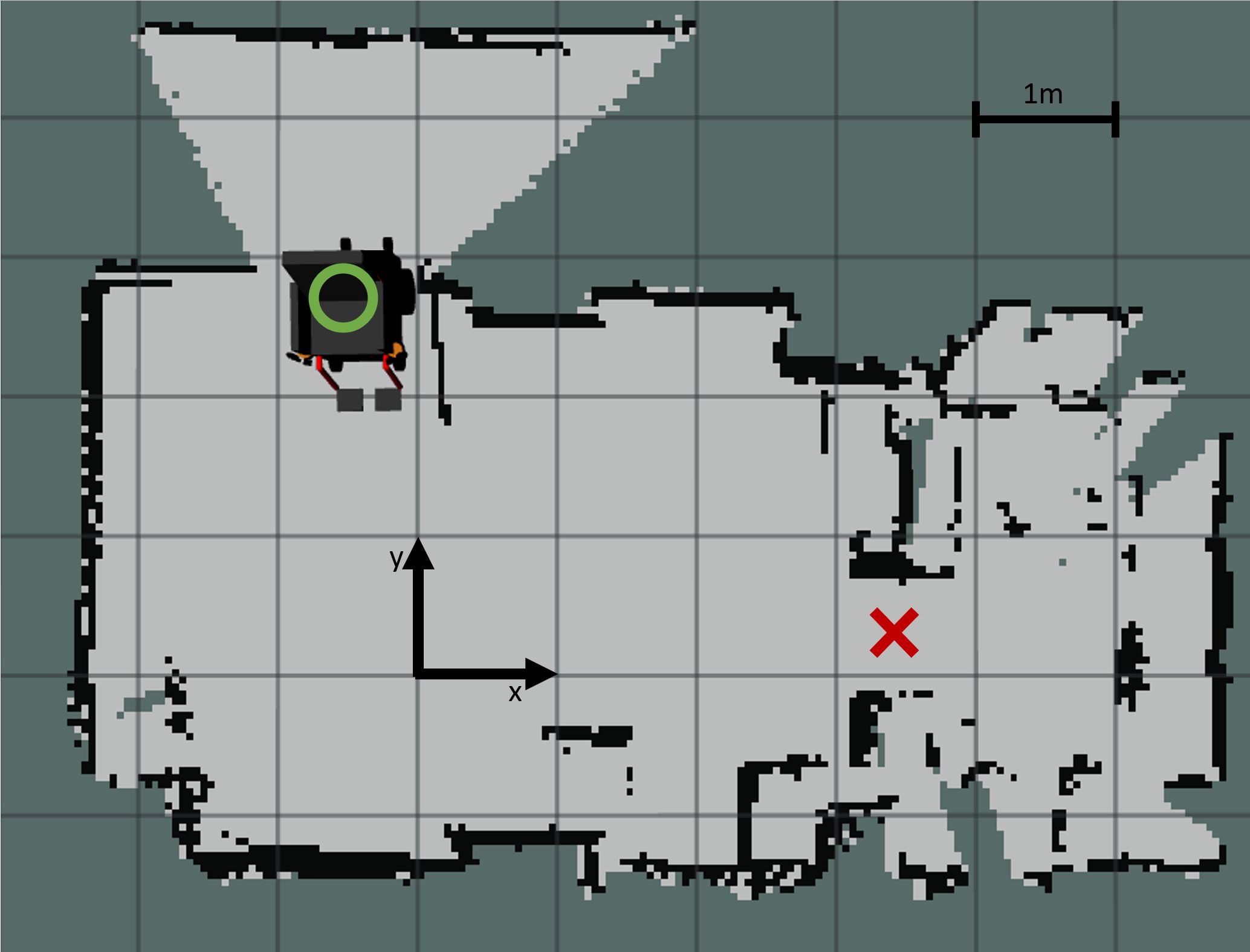}\\
    \caption{A cartographed map of the experimental environment using the RTAB-Map library. The start and end positions are located in the upper left corner (green circle) and along the right side (red cross), respectively.}\label{fig:lab}
    \end{center}
  \end{figure}

The task requires participants to push the \gls{ipw} from a designated start position to a predefined end position. The start position is located within a doorframe (upper left corner in Fig.~\ref{fig:lab}), while the end position is situated just beyond an opening leading to an adjacent lab room (on the right side in Fig.~\ref{fig:lab}). Participants control the \gls{ipw} using the extended handles (see Fig.~\ref{fig:ipw}) and are instructed to perform the movement smoothly and swiftly. After having reached the end position, the \gls{ipw} is to be pulled back to the start position.

\subsection{Experimental procedure}
\begin{figure}
  \begin{center}
  \includegraphics[width=0.5\columnwidth]{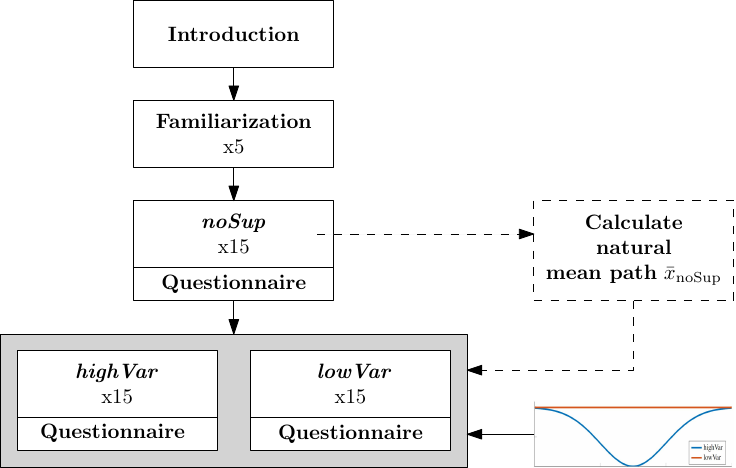}\\
  \caption{Study procedure. Solid lines depict the participant's workflow. Dashed lines depict supervisor's tasks.}\label{fig:procedure}
  \end{center}
\end{figure}

The study procedure consists of five sequential phases, resulting in a total duration of approx. $\SI{50}{mins}$ per participant.
An overview of the procedure is shown in Fig.~\ref{fig:procedure}.

Before beginning the experiment, participants receive an introduction, which includes information on the study's general objective, data privacy regulations, safety instructions, and a demonstration of the task procedure. 
The difference between the modes is not disclosed.

To familiarize participants with the \gls{ipw} and task mechanics, they first perform five familiarization trials in \mNosup mode. 
This phase allows them to become accustomed to the \gls{ipw} behavior.

Following familiarization, participants complete three experimental conditions in a within-subjects design:
\begin{itemize}
  \item \textbf{No Support} (\mNosup) mode: Participants perform $15$ repetitions of the task without any automation assistance.
  \item \textbf{Low Variability Assistance} (\mLowvar) mode: Participants perform $15$ repetitions with continuous automation assistance.
  \item \textbf{High Variability Assistance} (\mHighvar) mode: Participants perform $15$ repetitions with variability-preserving assistance.

\end{itemize}

All participants begin with mode \mNosup. The subsequent order of the \mLowvar and \mHighvar conditions is pseudo randomized to minimize order effects. The participants do not receive knowledge regarding the order and behavior of the modes. 

After completing each of the three conditions, participants fill out self-assessment questionnaires to evaluate their experience under each mode.

The experimental protocol was approved by the Research Ethics Board of the Université de Sherbrooke (CÉR Lettres et sciences humaines) under the application no. 2024-4639. The methods were carried out in accordance with the relevant guidelines and regulations.  Written informed consent was obtained from all participants before the study commenced.

\subsection{Data processing} \label{sec:processing}

 Before analyzing the recorded position data, a preprocessing pipeline was applied. Since start and end time were not defined, all trajectories were cropped at predefined spatial boundaries to ensure that only movement data between start point and end point is analyzed. Specifically, data points where the \gls{ipw}'s position exceeded $p_x > \SI{2.85}{m}$ or $p_y > \SI{2.35}{m}$ were discarded. These thresholds describe the lines behind which the IPW is located within the doorframe. 

 To maintain data quality, any trial was excluded from further analysis if the localization algorithm failed to provide an updated map transformation for a continuous travel distance exceeding $\SI{50}{cm}$. This selection ensured that only accurately localized trajectories were included in the final analysis. Regarding the study presented in this paper, $13$ out of $945$ trials were excluded. 

 To standardize path representation and allow for comparison across trials, spline interpolation was applied. All trajectories are resampled to a uniform length of $N = 300$ evenly spaced data points, identical to the human natural mean movement processing in Sec.~\ref{sec:humanMovement}. This step ensured that variability measures and comparisons were not influenced by differences in velocity.

 Subsequently, a mean path $\xMean = [\bar{\bm{p}}_{x,k} \;\bar{\bm{p}}_{y,k} ]^T $ was computed for each participant for each mode by averaging their recorded paths across all valid forward repetitions $c$, for each position $k$ along the path. The movement error $e_k$ at step $k$ was quantified by calculating the Euclidean distance between each individual trajectory and the corresponding mode's mean path: 
 \begin{align}
 e_k = \sqrt{ (p_{x,k}-\bar{p}_{x,k})^2 + (p_{y,k}-\bar{p}_{y,k})^2}.
 \end{align}
 
 The mode's variability was calculated based on the movement error: 
 \begin{align}
 \mathrm{cov}(e_k) = \frac{1}{c-1} \sum_{i=1}^{c} |e_k|^2 . 
 \end{align}

 The input measurements $\uHTrans$, $\uHRot$ and $\uARot$ were cropped to the same start and end points as the position data. The measurements were then also sampled using spline interpolation to a uniform length of $N = 300$ evenly timed data points.

\subsection{Objective measures} \label{sec:objMeasures}

Objective measures were extracted to quantitatively assess the effect of the automation modes on both task performance and cooperative behavior. The following key metrics were computed:

\begin{itemize}
    \item \textbf{Task-Irrelevant Position Variability (Max. Var.):} The variance of errors $\cov{e_k}$ serves as a measure of path variability. The maximum value along the path was taken as the task-irrelevant position variability value: $ \mathrm{max}(\cov{e})$.
    \item \textbf{Endpoint Error and Endpoint Variability:} To evaluate task-relevant performance, the mean over the error $ \bm{e}_k$  
    was calculated for the last $25$ data points, representing approximately the last $\SI{50}{cm}$ of the path which spans the length of the doorframe within which the goal position is located. The variance of this endpoint accuracy across repetitions was computed to capture endpoint variability.
    \item \textbf{Work Human:} The cumulative rotational input of the human participant and the automation were computed to assess the distribution of effort in cooperative control. The human work is defined as the sum of the human rotational input over the normalized trial: $W_H = \sum_{k=1}^{N} u_{H,\mathrm{rot},k}$. Since the input was normalized to match the \gls{ipw}'s admissible set, the work value is dimensionless.
\end{itemize}

These objective measures provide insight into how the different automation modes influence movement variability, accuracy, and the balance of control between human and machine.

\subsection{Subjective measures}\label{sec:experience}

In addition to objective performance metrics, subjective measures were collected to assess participants’ perceived experience under each experimental condition. Three standardized questionnaires were used:

\begin{itemize}
    \item \textbf{Sense of Agency (SoA):} A short-form version consisting of 13 items \cite{Tapal.2017}, evaluating participants' perceived agency over the \gls{ipw} movements during the task. The 7-point Likert scale ranged from 1 (strongly disagree) to 7 (strongly agree), with a higher overall value indicating a stronger perceived sense of agency.
    \item \textbf{Questionnaire of the Evaluation of Physical Assistive Devices (QUEAD2):} A 16-item version of the QUEAD~\cite{Schmidtler.2017}, which measures multiple dimensions of user experience, including perceived usefulness and perceived ease of use. The 7-point Likert scale ranged from 1 (strongly disagree) to 7 (strongly agree), with a higher overall value indicating a more positive user experience.
    \item \textbf{User Experience Questionnaire (UEQ):} A short-form version consisting of 8 items~\cite{Laugwitz.2008}, measuring overall user experience, particularly regarding usability and satisfaction. The 7-point Likert scale asks users to express their agreement towards one attribute in pairs of opposing attributes, e.g. "obstructive" vs. "supportive". The resulting scale ranges from -3 to 3, with a higher overall value indicating a more positive user experience.
\end{itemize}

These subjective measures allowed for a comprehensive evaluation of how different levels of automation and variability preservation impacted participants' sense of control, acceptance of the assistance system, and overall interaction experience.

\section{Results}

A total of 21 participants took part in the study, of which three had to be excluded from analysis due to faulty data logging. Out of these, four were female and 14 male; 15 participants were in their 20s, three in their 30s. All participants provided written consent concerning their participation. 

To analyze the effect of the automation mode on various measures, we use a repeated-measures one-way ANOVA. A Greenhouse-Geisser (GG) correction is applied, if the sphericity assumption does not hold. A t-test is used for pairwise comparison. Effects are considered marginally significant at $p < 0.1$, significant at $p < 0.05$, and highly significant at $p < 0.01$.

\subsection{Cooperative performance} \label{sec:results:objective}

\begin{figure*}[tbp]
  \centering

  \begin{subfigure}[t]{0.98\textwidth}
    \centering
    Subject A
    \resizebox{0.98\textwidth}{!}{
      \input{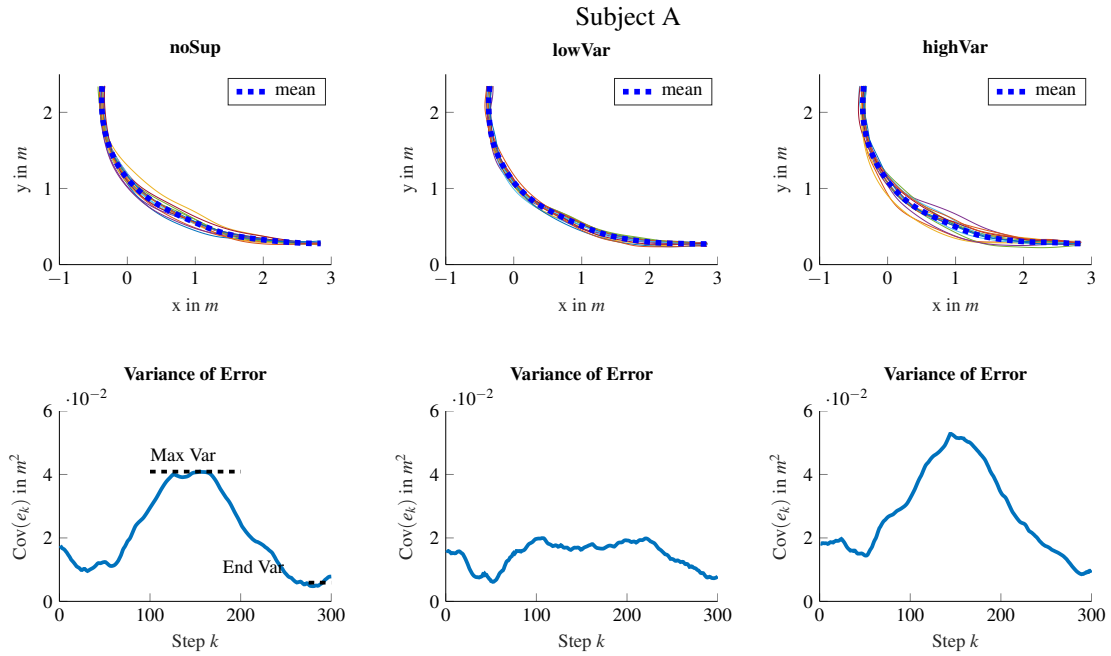}
      }

    \caption{Forward movements of one subject in all three modes. Top: 15 repetitions and their mean. Bottom: Position variance over all repetitions. Maximum and endpoint variance are depicted exemplary in \mNosup. }\label{fig:sub5}

  \end{subfigure}
  \hfill

  \vspace{10mm}

\begin{subfigure}[t]{0.98\textwidth}
    \centering
    All Subjects
    \resizebox{0.98\textwidth}{!}{
      \input{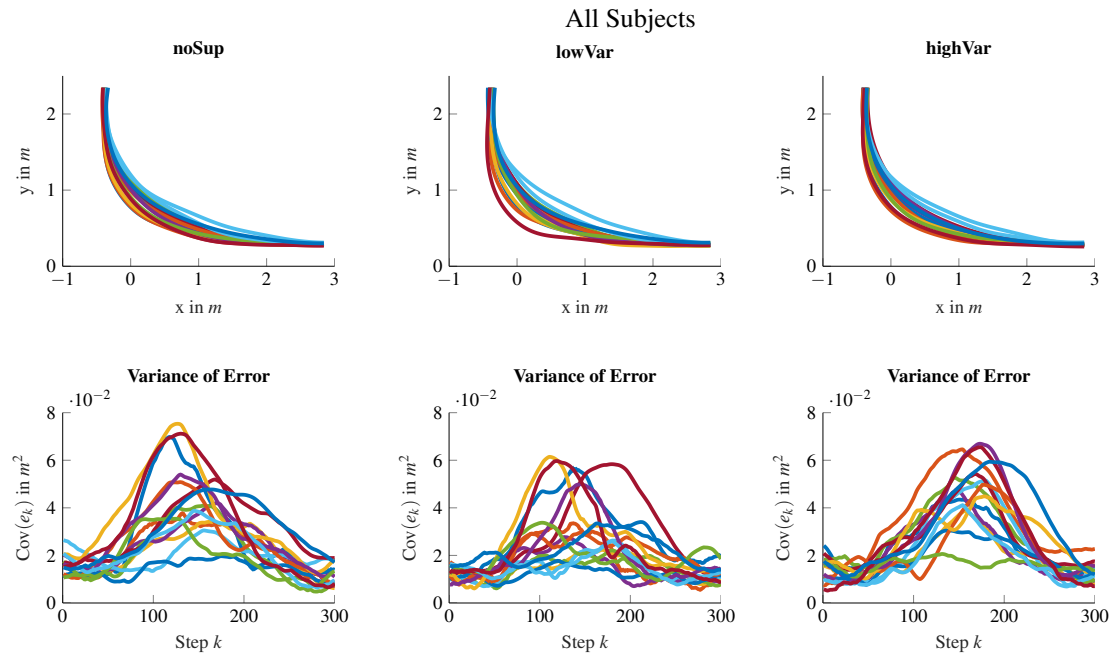}
      }
    \caption{Forward movement behavior of all subjects. Top: Mean paths of every subject. Bottom: Position variance of every subject. }\label{fig:allSub}
  \end{subfigure}
\caption{Behavior in all three modes: Subject~A (a) and all subjects (b). }
\end{figure*}

The trajectories and their variance of one subject performing 15~repetitions in all three modes are depicted in Fig.~\ref{fig:sub5}. In the \mNosup mode, a bell-shaped variability pattern can be observed: low variability at the start and end positions (approx. $\SI{0.02}{m^2}$ and $\SI{0.01}{m^2}$, respectively) and a comparably high maximum variability of approximately $\SI{0.04}{m^2}$ in the midsection between start and end point. The \mLowvar mode results in a consistently low variability along the entire path, with values remaining below $ \SI{0.02}{m^2}$. In contrast, the \mHighvar mode exhibits a pattern similar to \mNosup, showing reduced variability at start and end positions compared to elevated variability in the midway section, peaking around $\SI{0.05}{m^2}$.

\begin{figure}[btp]
  \centering
  \resizebox{0.98\columnwidth}{!}{
            \input{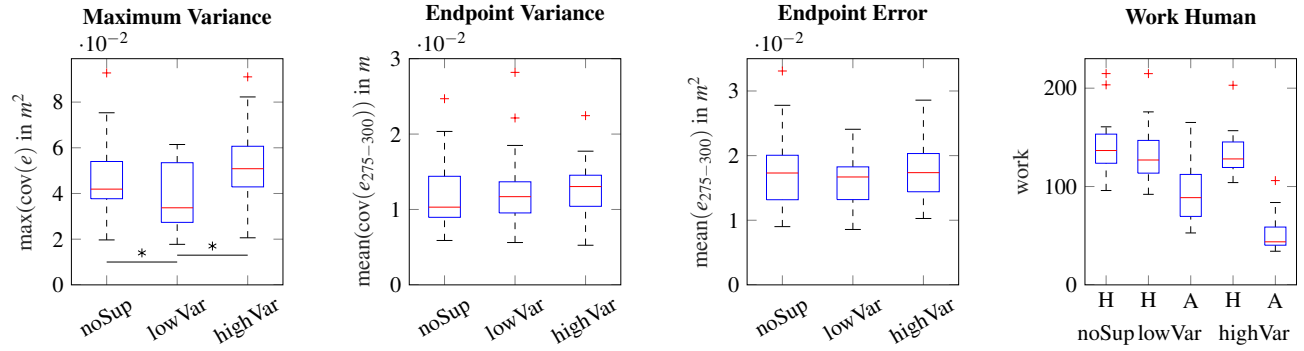}
      }
  \caption{Comparison of maximum and endpoint variability, error, and human work between the three modes. Significance levels: $^{\#}p < 0.1$, $^{*}p < 0.05$, $^{**}p < 0.01$.}
  \label{fig:boxplotObjective}
\end{figure}

The behavior across all subjects is depicted in Fig.~\ref{fig:allSub}, quantitatively listed in Table~\ref{table:statAnalysis}  and summarized in Fig.~\ref{fig:boxplotObjective}.

Notably, 12 out of 18 participants exhibited a reduced maximum variability in mode \mLowvar compared to both \mNosup and \mHighvar modes:
\begin{align*}
  \mathrm{max}\big(\cov{e}\big)_\mathrm{highVar} &> \mathrm{max}\big(\cov{e}\big)_\mathrm{lowVar}, \qquad \mathrm{and} \\
  \mathrm{max}\big(\cov{e}\big)_\mathrm{noSup} &> \mathrm{max}\big(\cov{e}\big)_\mathrm{lowVar}.
\end{align*}

Statistical analysis using repeated-measures ANOVA revealed a significant effect of the automation mode on maximum variability, with $F(2, 34) = 4.06, p = 0.026$ (sphericity assumption was met). Pairwise comparisons indicated that maximum variability was significantly higher in both \mNosup ($M = 0.049\,\mathrm{m}^2$, $p = 0.030$) and \mHighvar ($M = 0.052\,\mathrm{m}^2$, $p = 0.045$) compared to \mLowvar ($M = 0.041\,\mathrm{m}^2$), while \mNosup and \mHighvar did not differ significantly ($p = 1.000$).

No significant effect of automation mode was found on the mean endpoint error or endpoint variance, as desired. 

Regarding human work, the mean values were \mNosup ($M = 143.7$), \mLowvar ($M = 131.9$), and \mHighvar ($M = 135.6$). A significant overall effect of the automation mode on human work was found ($F(1.44, 24.49) = 4.01, p = 0.043$, Greenhouse--Geisser corrected); however, no individual pairwise comparison reached significance (all $p \geq 0.129$).


\begin{table*}[h]
\caption{Mean $\pm$ standard deviation (top), repeated-measures ANOVA results with partial eta-squared $\eta_p^2$ (middle), and pairwise comparisons with Cohen's $d_z$ effect sizes (bottom).
Greenhouse--Geisser correction applied where sphericity was violated (Endpoint Error, Work Human).
Statistical significances: $^{\#}p < 0.1$, $^{*}p < 0.05$, $^{**}p < 0.01$.}
\label{table:statAnalysis}
\footnotesize
\begin{tabular*}{\textwidth}{@{\extracolsep\fill}lccccccccc}
\toprule
 &
\begin{tabular}[c]{@{}c@{}}Max.\\Var.\\$[10^{-2}\,\mathrm{m}^2]$\end{tabular} &
\begin{tabular}[c]{@{}c@{}}Endpt.\\Var.\\$[10^{-2}\,\mathrm{m}^2]$\end{tabular} &
\begin{tabular}[c]{@{}c@{}}Endpt.\\Error\\$[10^{-2}\,\mathrm{m}]$\end{tabular} &
\begin{tabular}[c]{@{}c@{}}Work\\Human\end{tabular} &
\begin{tabular}[c]{@{}c@{}}Sense of\\Agency\end{tabular} &
Usability &
\begin{tabular}[c]{@{}c@{}}Perceived\\Usefulness\end{tabular} &
\begin{tabular}[c]{@{}c@{}}Perceived\\Ease of Use\end{tabular} &
\begin{tabular}[c]{@{}c@{}}User\\Experience\end{tabular} \\
\midrule
\mNosup
& $4.92 \pm 1.82$ & $1.24 \pm 0.52$ & $1.84 \pm 0.60$ & $143.72 \pm 28.54$
& $5.69 \pm 0.57$ & $4.55 \pm 0.56$ & $4.67 \pm 0.95$ & $4.96 \pm 0.93$ & $0.41 \pm 1.06$ \\
\mLowvar
& $4.06 \pm 1.52$ & $1.30 \pm 0.54$ & $1.67 \pm 0.44$ & $131.90 \pm 27.29$
& $4.81 \pm 0.93$ & $4.62 \pm 0.67$ & $5.08 \pm 0.79$ & $5.01 \pm 0.90$ & $1.17 \pm 0.84$ \\
\mHighvar
& $5.21 \pm 1.70$ & $1.22 \pm 0.32$ & $1.75 \pm 0.48$ & $135.56 \pm 21.66$
& $5.48 \pm 0.75$ & $4.79 \pm 0.67$ & $5.33 \pm 0.93$ & $5.44 \pm 1.14$ & $1.10 \pm 1.00$ \\
\midrule
$\mathrm{df}_1,\,\mathrm{df}_2$
& 2,\,34 & 2,\,34 & 1.49,\,25.5 & 1.44,\,24.5 & 2,\,34 & 2,\,34 & 2,\,34 & 2,\,34 & 2,\,34 \\
$F$
& 4.06 & 0.26 & 1.33 & 4.01 & 10.96 & 1.06 & 5.61 & 3.05 & 9.32 \\
$\eta_p^2$
& 0.19 & 0.02 & 0.07 & 0.19 & 0.39 & 0.06 & 0.25 & 0.15 & 0.35 \\
$p$
& $\mathbf{0.026^{*}}$ & 0.770 & 0.275 & $\mathbf{0.043^{*}}$
& $\mathbf{<0.001^{**}}$ & 0.359 & $\mathbf{0.008^{**}}$ & $\mathbf{0.060^{\#}}$ & $\mathbf{<0.001^{**}}$ \\
\midrule
Comparison & & & & & & & & & \\
\midrule
\mNosup--\mLowvar
& $\mathbf{0.030^{*}}$ & 1.000 & 0.399 & 0.129
& $\mathbf{0.002^{**}}$ & 1.000 & 0.188 & 1.000 & $\mathbf{0.005^{**}}$ \\
\hspace{1em}$d_z$
& 0.68 & $-0.11$ & 0.37 & 0.52
& 0.97 & $-0.08$ & $-0.47$ & $-0.06$ & $-0.87$ \\
\mNosup--\mHighvar
& 1.000 & 1.000 & 1.000 & 0.160
& 0.692 & 0.302 & $\mathbf{0.010^{*}}$ & $\mathbf{0.069^{\#}}$ & $\mathbf{0.007^{**}}$ \\
\hspace{1em}$d_z$
& $-0.13$ & 0.04 & 0.17 & 0.49
& 0.29 & $-0.41$ & $-0.80$ & $-0.59$ & $-0.84$ \\
\mLowvar--\mHighvar
& $\mathbf{0.045^{*}}$ & 1.000 & 0.809 & 0.798
& $\mathbf{0.013^{*}}$ & 1.000 & 0.672 & 0.270 & 1.000 \\
\hspace{1em}$d_z$
& $-0.64$ & 0.17 & $-0.27$ & $-0.27$
& $-0.77$ & $-0.23$ & $-0.30$ & $-0.42$ & 0.09 \\
\bottomrule
\end{tabular*}
\end{table*}

\subsection{Human interaction experience}

\begin{figure*}[btp]
  \centering
  \resizebox{.98\textwidth}{!}{
      \input{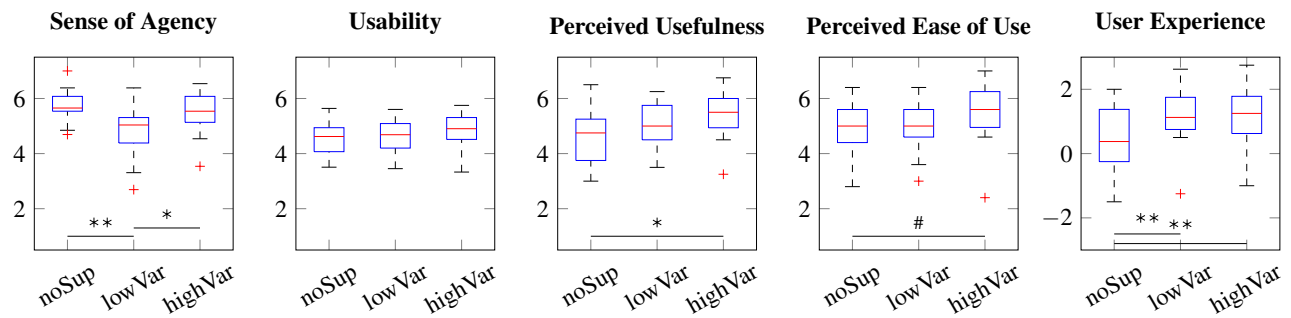}
      }
  \caption{Subjective experience of sense of agency, usability (including the subscales perceived usefulness and ease of use) and user experience, for all three modes. All scales, except user experience, range from 1 to 7. User experience ranges from -3 to 3. Significance levels: $^{\#}p < 0.1$, $^{*}p < 0.05$, $^{**}p < 0.01$.
}
  \label{fig:boxplotSubjective}
\end{figure*}

The impact of the different automation modes on subjective human experience was evaluated using three previously introduced questionnaires. The results of these assessments are summarized in Fig.~\ref{fig:boxplotSubjective}.

Analysis of the sense of agency ratings revealed a significant effect of the automation mode (sphericity assumption was met). The repeated-measures ANOVA indicated a significant effect of the mode ($F(2,34) = 10.96, p < 0.001$) on the perceived sense of agency. Pairwise comparisons showed that the rating was significantly higher in the \mNosup ($M = 5.69$, $p = 0.002$) and \mHighvar ($M = 5.48$, $p = 0.013$) conditions compared to \mLowvar ($M = 4.81$).
No significant difference in sense of agency was found between \mNosup and \mHighvar.

For the usability questionnaire, which evaluates perceived usefulness, ease of use, and overall acceptance of the assistive device, no significant effect of the automation mode on the overall score was found. Further analysis of the subscales indicated that perceived usefulness showed a significant effect of the mode ($F(2,34) = 5.61, p = 0.008$), with \mHighvar ($M = 5.33$) rated significantly higher than \mNosup ($M = 4.67$, $p = 0.010$). Perceived ease of use showed a marginally significant effect ($F(2,34) = 3.05, p = 0.060$), with \mHighvar ($M = 5.44$) rated marginally higher than \mNosup ($M = 4.96$, $p = 0.069$).

Across all usability subscales, \mHighvar consistently received higher ratings than \mLowvar, though these differences did not reach statistical significance.

Regarding the user experience questionnaire, a significant effect of the mode was observed (sphericity assumption was met). ANOVA analysis yielded $F(2,34) = 9.32, p < 0.001$. Pairwise comparisons showed that both \mLowvar ($M = 1.17$, $p = 0.005$) and \mHighvar ($M = 1.10$, $p = 0.007$) were rated significantly higher than \mNosup ($M = 0.41$). No significant difference was detected between \mHighvar and \mLowvar.

In summary, \mHighvar and \mNosup exhibited comparable levels of sense of agency, both significantly higher than \mLowvar, as intended by the study design. While no significant difference was observed in overall usability, the automation mode significantly affected perceived usefulness and user experience, and showed a marginal effect on perceived ease of use. Specifically, \mHighvar received significantly higher ratings than \mNosup for perceived usefulness and user experience, and marginally higher ratings for perceived ease of use. In addition, \mLowvar also received significantly higher user experience ratings than \mNosup.

\section{Discussion}

The results of our study offer valuable insights into the role of preserving natural human locomotion variability in HRI, regarding objective and subjective implications. 

As hypothesized, the analysis of cooperative performance in Sec.~\ref{sec:results:objective} demonstrated that the automation mode significantly influences task-irrelevant variability. Specifically, the \mLowvar mode effectively reduced variability throughout the entire movement, whereas the \mHighvar mode preserved higher variability in task-irrelevant regions, closely resembling the natural bell-shaped variability pattern observed in the \mNosup condition. This validates the automation modes and accepts the first hypothesis \textit{H1}, confirming that the designed variability modulation in \mHighvar mode enables participants to retain their natural movement variability while still benefiting from assistance in task-relevant areas.

The results also showed that the \mHighvar mode did not significantly differ from \mNosup in terms of maximum variability, indicating that the assistance system successfully preserved natural variability patterns. Moreover, no significant differences in task-relevant performance measures---such as endpoint error and endpoint variance---were observed between the modes, supporting hypothesis \textit{H2}. Taken together, these findings suggest that respecting natural variability can be achieved without compromising task-relevant performance.
From a system design perspective, this suggests that respecting natural variability can be achieved without negatively impacting overall task success.

While our analyses revealed statistically significant differences in variability between modes, it is also important to consider whether these differences are perceptible to users and thus practically meaningful. The observed difference in maximum variability between \mHighvar and \mLowvar, which is statistically significant ($p=0.045$), corresponds to an average increase of $0.012\,\mathrm{m}^2$. Given the scale of the task and the overall trajectory length, this magnitude of variability likely reflects a small but noticeable divergence in path behavior, which may contribute to the subjective experience of control. Although perceptual thresholds for trajectory variability in HRI are not yet formally established, the observed increase represents a divergence large enough to manifest as a visibly broader trajectory spread, which aligns with participants’ subjective reports: Some participants mentioned that in \mLowvar, the IPW moved like it was slightly constrained on rails, while in \mHighvar, the IPW felt lighter and easier to control.
The differences in endpoint error, statistically tested and shown to be not significantly different, were very small ($< \SI{2}{mm} $) and thus unlikely to be perceived by participants.

Regarding inter-subject variability, we can observe that when moving naturally in \mNosup mode, the mean paths and variability shapes vary between the subjects. This shows the need for personalized assistance systems. While this study respected the individual mean behavior of each participant by providing an assistance tailored to the individual mean path identified from the \mNosup mode, the implemented variability patterns were static for all subjects.


Regarding subjective experience, the results provide insights into how variability modulation affects interaction experience. 

Specifically, the sense of agency in the \mLowvar mode was significantly lower than in the other two modes, wile \mHighvar exhibited comparable sense-of-agency ratings to \mNosup. 
This suggests that excessive---and specifically variability-restraining---assistance can diminish user's perceived ownership of the overall action, whereas assistance that selectively supports task-relevant aspects of the movement can be integrated into the user's own sense of agency.

These findings are consistent with motor prediction models, which emphasize that the sense of agency depends not only on physical effort but also on the congruence between intended actions and perceived outcomes~\cite{Wen.2015, Blakemore.2002}. One possible explanation for the reduced sense of agency observed in the \mLowvar mode is that continuous system interventions introduce a mismatch with the user's intended movement variability. In contrast, variability-preserving control may help maintain a higher sense of agency by respecting users' natural movement structure. 

Given the required trade-off between automation support and user autonomy~\cite{Kille.2025}, the present results tentatively point toward variability-preserving control designs as a promising design direction.
By providing assistance selectively---enhancing support in task-relevant aspects of the movement where it aligns with user intentions, while allowing natural variability in task-irrelevant dimensions---such approaches may help preserve the sense of agency without compromising task performance or perceived autonomy.

The observed effects may be particularly noteworthy given that the reference trajectories were personalized. For both assistance modes, the mean reference path was computed individually for each participant based on their natural trajectories in the \mNosup condition. Consequently, the reference trajectory underlying the \mLowvar mode was already closely aligned with each participant’s inherent movement behavior.

It is therefore plausible that this personalization elevated the perceived sense of agency in the \mLowvar condition, as participants may have experienced the assistance as reinforcing their own natural trajectories rather than imposing external control. While this design choice is beneficial from a user-centered perspective, it may also have attenuated the observable differences between the two assisted modes, thereby reducing the statistical contrast between \mLowvar and \mHighvar. Using a non-personalized reference trajectory across participants would likely further reduce the sense of agency in \mLowvar.

While no significant differences were observed in overall usability scores, the \mHighvar mode consistently received higher ratings than both \mLowvar and \mNosup across all usability subscales, with perceived usefulness reaching statistical significance ($p = 0.010$) and perceived ease of use reaching marginal significance ($p = 0.069$) in favor of \mHighvar over \mNosup. Their consistent direction aligns with the notion that retaining natural movement variability may positively influence perceived usefulness and ease of use.

Taken together with the sense-of-agency results, these findings provide partial and exploratory support for the third hypothesis (\textit{H3}). Specifically, while \mHighvar significantly increased the sense of agency compared to \mLowvar, and significantly outperformed \mNosup on perceived usefulness and marginally on perceived ease of use, neither usability subscale differed significantly between the two assisted modes. This suggests that variability-preserving assistance improves perceived usability relative to no support, but its advantage over conventional assistance remains below statistical significance---possibly reflecting that the personalized reference trajectory already partially aligned \mLowvar with individual user behavior.

One such factor may be the personalization of reference trajectories. By grounding both \mLowvar and \mHighvar in each participant’s natural behavior observed in the \mNosup condition, the assistance in \mLowvar may have appeared more aligned with user intentions than a generic reference would have, thereby elevating its perceived agency and usability. Consequently, while the usability results for perceived usefulness are now sufficiently robust ($p = 0.010$), the absence of a statistically significant difference between the two assisted modes means that the subjective results should still be interpreted cautiously with respect to hypothesis \textit{H3}.

The user experience questionnaire revealed a comparable pattern. Both the \mLowvar and \mHighvar modes received similar user experience ratings, which were significantly higher than those of the \mNosup condition. Given that the two assisted modes did not differ significantly in user experience despite differences in sense of agency, user experience ratings in this study cannot be explained by agency alone. Instead, these results suggest that additional factors—such as perceived novelty or initial engagement with automation—may have contributed more strongly to user experience ratings than agency or usability measures.


The findings of this study underline a key trade-off in shared control system design. While high levels of assistance and constraint may produce more uniform and stable behavior, they risk undermining the user's sense of control and thereby potentially affect the overall acceptance and satisfaction with the system. Our results suggest that preserving natural variability---particularly in task-irrelevant areas---may offer a promising approach to balancing technical performance with a positive user experience.

This has potential implications for the development of assistive robots and rehabilitation devices. Systems that selectively support users while respecting inherent human variability could improve both functional outcomes and long-term user engagement, acceptance, and satisfaction. This aligns with current research in HMI that analyzes various psychological states during interaction \cite{Orlando.2025, VanDerWoerdt.2019, Zanatto.2021a} or examines the effect of novel measures such as human-robot fluency on interaction experience \cite{Paliga.2021}. Taken together, the objective and subjective results suggest that natural variability can be preserved without loss of task-relevant performance. While our findings highlight the potential of variability-aware shared control as a design principle, further studies are needed to establish its generalizability and long-term impact.


Building upon these findings, future work should explore strategies to personalize assistance including the stochastic nature of human behavior. Further research could analyze, whether the incorporation of personalized variability patterns, instead of a generic one, into the control design lead to enhanced user experience. Since subjects exhibited various levels of variability in \mNosup, it might be of benefit to reflect their natural variability behavior within the automation. It may also be of interest, how non-personalized mean trajectories as basis for assistance may have an effect on the modulation of interaction experience. A study design that explicitly allows high and low variability without additionally introducing personalization measures which might mitigate sense of agency differences would shed light onto this question. Additionally, longitudinal studies could investigate how variability-preserving assistance affects user engagement, learning, and adaptation over extended periods. While the experience of novelty, which potentially led to slightly elevated ratings in the user experience, might fade over time, elevated sense of agency has been shown to increase user engagement long-term~\cite{Haggard.2012}. Further exploration of physiological or cognitive measures in relation to movement variability could also provide deeper insights into optimizing user experience in HRI. Future work should also investigate perceptual thresholds of variability in HRI, 
to better establish which magnitudes of variability differences are meaningfully experienced by users and which remain below subjective detection. This would also aid in potentially differentiating between groups of users that exhibit high vs. low natural variability, and whether they benefit differently from variability-preserving assistance.

An important limitation of the present study is that task-relevant and task-irrelevant areas were defined spatially along a point-to-point locomotion trajectory. 
While this framing is intuitive for wheelchair navigation, it restricts the generalizability of the approach. 
In more complex systems such as robot arms or collaborative manipulation  tasks, task-relevancy may need to be identified differently: for example, in terms of dimensions of the state space, temporal phases of a movement, or along manifolds such as those described by the uncontrolled manifold hypothesis. 
Distribution-based approaches, such as ProMPs as used by \cite{Maeda.2017}, could provide a principled way to identify high- and low-variance dimensions and to adapt variability-preserving control accordingly. 
Incorporating such representations represents a promising direction for future work.

However, extending the applicability of variability-aware shared control to a broader range of HRI tasks will require the development of additional variability-respecting control methods. 
In the present work, automation was implemented as a path-following controller with variable gain, limited to the rotational dimension of maneuvering a wheelchair. 
While this design fulfilled the aim of the study, it does not directly generalize to tasks beyond locomotion. 
In collaborative scenarios involving tools or workpieces, incorporating natural variability into commonly used impedance control designs may represent a promising direction for future research.

One final limitation of this study is the relatively small sample size (N = 18), which may have limited the statistical power to detect subtle effects in subjective ratings. 
While trends in the data suggest potential benefits of variability-preserving assistance, these should be interpreted with caution. 
Future studies with larger and more diverse participant groups are needed to validate these findings and confirm their generalizability. Additionally, wile the task was intentionally constrained to isolate variability effects, future work should examine whether similar patterns emerge in more ecologically complex HRI scenarios.

\section{Conclusion}

This paper investigated the relevance of natural human locomotion variability in HRI. We proposed a shared control framework that selectively preserves human inherent variability by providing assistance only in task-relevant areas while allowing natural variability in task-irrelevant regions. An IPW was adapted to be used with our framework. Through a user study involving interaction with the IPW under different automation modes, we evaluated the impact of preserving variability on both task performance and interaction experience.

Our results demonstrated that the variability-preserving assistance mode successfully retained natural movement variability without negatively affecting task-relevant performance measures. With respect to subjective assessments, we found that continuous assistance (\mLowvar) significantly diminished participants' sense of agency compared to the no-support baseline and compared to the variability-preserving mode (\mHighvar). Additionally, \mHighvar significantly outperformed \mNosup on perceived usefulness ($p = 0.010$) and showed a marginal effect on perceived ease of use ($p = 0.069$), while neither subscale differed significantly between the two assisted modes. 

Overall, the main contribution of this work lies in demonstrating through an empirical user study that variability-preserving shared control can retain natural variability without compromising task-relevant performance, and in highlighting the role of subjective measures such as sense of agency for evaluating shared control strategies. These findings underscore the importance of accounting for natural movement variability when designing human-centric shared control systems. By respecting inherent human behavior, variability-preserving strategies may foster user autonomy and acceptance without compromising task success.

\bibliography{bib}

\section*{Acknowledgements}

\subsection*{Funding}
Partial financial support was received from MITACS through the Globalink Research Award under Grant Agreement No. IT38656.

\subsection*{Ethics approval and consent to participate}
The experimental protocol was approved by the Research Ethics Board of the Université de Sherbrooke (CÉR Lettres et sciences humaines) under the application no. 2024-4639. Written informed consent was obtained from all participants before the study commenced.

\section*{Author contributions statement}

S.K., A.M.P., B.V., and S.H. conceived and designed the work. S.K. and A.M.P. acquired the data. S.K. analysed and interpreted the data. S.K. drafted the manuscript. S.K., A.M.P., B.V., and S.H. substantively revised the manuscript. All authors reviewed the manuscript.

\section*{Additional information}

\subsection*{Competing interests}
The authors have no relevant financial or non-financial interests to disclose.

\end{document}